\documentclass[a4paper,UKenglish,cleveref, autoref, thm-restate]{lipics-v2021}

\title{Data Driven VRP: A Neural Network Model to Learn Hidden Preferences for VRP \footnote{Preprint version. Article accepted for publication at CP, 2021}}
\titlerunning{Data Driven VRP}

\author{Jayanta Mandi}{Data Analytics Laboratory, Vrije Universiteit Brussel, Belgium  \and \url{https://jayman91.github.io/}}{jayanta.mandi@vub.be}{https://orcid.org/0000-0001-8675-8178}{}

\author{Rocsildes Canoy}{Data Analytics Laboratory, Vrije Universiteit Brussel, Belgium  }{rocsildes.canoy@vub.be}{https://orcid.org/0000-0003-1810-082X}{}

\author{V\'ictor Bucarey}{Institute of Engineering Sciences, Universidad de O'Higgins, Rancagua, Chile. \and \url{https://sites.google.com/site/vbucarey/}}{victor.bucarey@uoh.cl}{https://orcid.org/0000-0002-3043-8404}{}

\author{Tias Guns}{Data Analytics Laboratory, Vrije Universiteit Brussel, Belgium \and Department of Computer Science, KU Leuven, Belgium \and \url{https://people.cs.kuleuven.be/~tias.guns/}}{tias.guns@vub.be}{[orcid]}{}

\authorrunning{Mandi et al.} 

\Copyright{Jayanta Mandi and Rocsildes Canoy and  V\'ictor Bucarey and Tias Guns} 



\category{} 
\relatedversion{} 

\funding{Research supported by the FWO Flanders project Data-driven logistics (FWO-S007318N)}

\acknowledgements{}
\nolinenumbers 

\hideLIPIcs  

\EventEditors{Laurent D. Michel}
\EventNoEds{1}
\EventLongTitle{27th International Conference on Principles and Practice of Constraint Programming (CP 2021)}
\EventShortTitle{CP 2021}
\EventAcronym{CP}
\EventYear{2021}
\EventDate{October 25--29, 2021}
\EventLocation{Montpellier, France (Virtual Conference)}
\EventLogo{}
\SeriesVolume{210}
\ArticleNo{23}
\usepackage[utf8]{inputenc}
\usepackage{graphicx,xcolor}
\usepackage{array, stmaryrd}
\usepackage{booktabs,makecell}
\usepackage[ruled,linesnumbered]{algorithm2e}

\newcommand\jay[1]{\textcolor{blue}{}}
\newcommand\rocs[1]{\textcolor{orange}{}}
\newcommand\tias[1]{\textcolor{teal}{}}
\newcommand\victor[1]{\textcolor{blue}{}}
\begin{document}

\maketitle
\begin{abstract}
The traditional Capacitated Vehicle Routing Problem (CVRP) minimizes the total distance of the routes under the capacity constraints of the vehicles. But more often, the objective involves multiple criteria including not only the total distance of the tour but also other factors such as travel costs, travel time, and fuel consumption.
Moreover, in reality, there are numerous implicit preferences ingrained in the minds of the route planners and the drivers. Drivers, for instance, have familiarity with certain neighborhoods and knowledge of the state of roads, and often consider the best places for rest and lunch breaks. This knowledge is difficult to formulate and balance when operational routing decisions have to be made. 

This motivates us to learn the implicit preferences from past solutions and to incorporate these learned preferences in the optimization process. These preferences are in the form of arc probabilities, i.e., the more preferred a route is, the higher is the joint probability. The novelty of this work is the use of a neural network model to estimate the arc probabilities, which allows for additional features and automatic parameter estimation. 
This first requires identifying suitable features, neural architectures and loss functions, taking into account that there is typically few data available. We investigate the difference with a prior weighted Markov counting approach, and study the applicability of neural networks in this setting.
\end{abstract}

\section{Introduction}
Although the Vehicle Routing Problem (VRP) and its many variants have been extensively studied in the literature, the `theoretical optimal' solution often does not meet the expectations of the route planners and the drivers. This is because in real-life operations, the acceptability of a route is dependent not only on distance, travel time or fuel consumption, which have been studied in the literature, but also on multiple factors which are difficult to put in the objective function. A study by \cite{ceikute2013routing} has revealed that local drivers prefer routes that are not optimal in terms of travel time or cost.
The drivers take into account several factors which are not in the objective function such as traffic congestion and availability of parking and fuel stations.
This highlights the necessity of \emph{preference-based} routing where the objective is to minimize the travel cost as perceived by the drivers and the planners. To put it another way, we can see it as maximizing the utility of the drivers and the planners. 

In this work, we propose VRP solutions which are acceptable to the route planners and the drivers.
We start from the setting studied in \cite{canoy2019vehicle}, which proposes the maximum likelihood routing. Maximum likelihood routing considers the transition probabilities between the stops as \emph{revealed preferences} of the drivers and the planners and finds the maximum utility route by maximizing the joint transition probabilities.
To estimate the transition probabilities, \cite{canoy2019vehicle} uses a Markov counting approach which enumerates the past solutions for each realized route. Their approach uses only the past solutions for probabilities estimation but cannot make use of contextual features such as day of week.
We extend their framework to use a neural network model to estimate the transition probabilities before finding the route by applying maximum likelihood routing. The motivation behind using the neural network model is to generate a better estimation of the cost vector by using historical as well as contextual information in the neural network model.


We start with a neural network model which is trained using both contextual information and past solutions. We also include the Markov prediction as a feature in the neural network and observe improvement in the solution quality. Finally, we choose a parsimonious architecture in order to avoid overfitting and with this we are able to outperform  \cite{canoy2019vehicle}.

\textbf{Contributions}
\begin{itemize}
    \item We formulate the challenge of neural network-based learning of hidden preferences from moderately sized data, in a way that is compatible with existing VRP solvers.
    \item We investigate different features and architectures for such a neural network, more specifically arc-based linear models combined into per-node probabilistic estimates.
    \item We investigate how we can combine the Markov model and neural network, e.g., by considering the Markov predictions as an input to the neural network model.
    \item We propose two loss functions that allow for gradient-descent learning: one based on standard multi-class losses and another based on decision-focused learning that incorporate the VRP solving into the loss function.
\end{itemize}

\section{Related Work} 




The VRP \cite{dantzig1959truck} has been studied with its many different variations.
Traditional VRP minimizes a tangible objective such as
 operational costs \cite{hu2009a}, travel time \cite{lecluyse2009vehicle}, fuel consumption or carbon emission \cite{xiao2012development, peng2009research}.
Although multiple aspects of the assignment schedules of the drivers such as route balancing \cite{lee1999study} have been studied, learning and optimizing drivers' preferences has recently received increasing attention.

The preferences of the drivers can be considered by including them in the objective function. This can be treated in a
multi-objective VRP \cite{jozefowiez2008multi} setting, such as forming an objective function as a weighted sum or finding the set of Pareto optimal solutions based on standard multi-objective evolutionary algorithms \cite{schaffer1985multiple}. However, the preferences of the drivers are implicit \cite{toledo2013decision} and in most cases, explicit formalization of these preferences is not possible in practice.

Authors in \cite{canoy2019vehicle} tackled the problem from a different perspective: they introduce a weighted Markov model to learn the preferences. This approach avoids the explicit specification of the preference constraints and the implicit sub-objectives. The Markovian model is built using preferences learned from past solutions, which the planners have constructed by modifying solutions given by off-the-self solvers. Contrary to their work, we use a neural network model to learn the drivers' preferences, allowing a more flexible and general framework.

Learning preferences for drivers have been focused mainly in the setting of one single origin and destination. TRIP \cite{letchner2006trip} leverages past GPS data to learn drivers' preferences by comparing the ratios of the drivers' travel time to the average travel time, and with that, it generates routes that mimic the ones chosen by the drivers. The approach in \cite{funke2016deducing} also deduces driving preferences from GPS traces and models them into the weights of a linear programming formulation, which is then optimized to generate new route suggestions.
The authors in \cite{guo2020context} are able to enhance the quality of the solutions by considering different routing preferences that vary depending on the contexts. 
While we do not provide an explicit representation of the preferences, we assume that the preferences can be expressed in terms of probabilities (or utilities) of the arcs in the graph. 

Decision-focused learning \cite{wilder2019,elmachtoub2021smart}, which combines gradient-based neural network and optimization into a single framework, has recently received much attention in operations research. In this setup, the outputs of the neural network are fed into the optimization module as one of the inputs. The novelty of this approach is that it trains the neural network model while considering the objective value in the optimization problem. Decision-focused learning of submodular optimization problems, zero-sum games and SAT problems have been studied in \cite{wilder2019, ling2018what, wang2019satnet} respectively. The approach proposed by \cite{Blackbox2020differentiation} also combines a neural network model with any given optimization oracle via  `implicit interpolation.'  Ours is the first work which uses this framework for learning preferences in the context of vehicle routing.


\section{Preliminaries}
\subsection{Problem Description}
In this work, we are interested in the route planning process of an actual small transportation company. The route planners in the company are responsible for organizing tours for a fleet of vehicles in order to deliver goods to the customers.  Although they use a commercial route optimization software to produce routes that are optimal in terms of route length and travel time, they are hardly satisfied with the solutions. Solutions have to be modified to come up with a tour which is acceptable to the planners, drivers, and other stakeholders. In this way, the planners are implicitly optimizing the utilities of all those involved.

One way to approach this problem is to explicitly define the set of objectives. However, it is nearly impossible to model such personal preferences. 
We observe that the planners start from past realized routes because they require minimal modifications compared to the `theoretical optimal' routes. Therefore, in a way, the past solutions capture the preferences (or the utilities) of the planners and the drivers.
Our objective in this work is to learn the latent preferences of the drivers and the route planners for vehicle routing using a neural network model and propose tours which are acceptable to the planners. More specifically, we will focus on learning preferences at the arc level. We consider the transition probabilities as \emph{revealed preferences}.  We use neural network to output the transition probabilities between every pair of nodes. 
The advantage of this formulation is that we can use the negative log probability in place of a traditional travel cost in any existing VRP solver.

\subparagraph*{Challenges.} A machine learning model learns from training instances. In our case with the company data, each instance is realized in a day. However, due to functional and operational reasons, tours are not organized each day. 
Putting that into perspective, it would take more than 6 months to collect only 180 training instances.
Consequently, we are not in a state to use a neural network model trained with thousands of training instances. Hence, we have to be particularly careful about using a neural network model with a large number of parameters, as such network is prone to overfitting with small data.

Another challenge in this case is that not all customers raise a demand request with equal frequency. In fact, some customers have daily requests, whereas others raise requests only once or twice in a month. Depending on which set of customers raise a request, there can be considerable changes in the tour. We also observe a weekly pattern in the tours, i.e., tours of one weekday are different from the other days but very similar to those from the same weekday of the previous weeks. Therefore, learning the weekly patterns from a limited number of weeks poses another challenge.


\subsection{Formalization}
We begin by formalizing the objective and the data structures.
Formally, on a given day $t$, $S^t$ is the set of stops to be served by a number $m^t$ of homogeneous vehicles. 
We represent  $S^t \doteq \{0,1,\ldots,n\}$, where 0 represents the depot, and the other nodes represent the customers. Let $A^t$ define the set of all arcs in $S^t$.

We call $\mathbf{x}^t$ a \textbf{routing} 
with respect to $S^t$ with $m^t$ homogeneous vehicles, if $\mathbf{x}^t$ contains a set of at most $m^t$ tours in $S^t$ with each tour starting from and ending at the depot $0$ and each node in $S^t$ is visited exactly once to satisfy its demand request. Additionally, a feasible routing should ensure that the total demand allocated to each vehicle does not exceed its capacity $Q$.
Let  ${\mathcal X}_{S^t}^{m^t}$ denote all feasible routings of $m^t$ vehicles over $S^t$.
The objective in standard CVRP is to minimize the total travel costs of the routing. 
We remark that the depot is fixed and always present but the set of stops $S^t$ changes from one day to another as not all customers raise a demand request each day. 

For learning the preferences from past data, we are given a dataset $\mathcal{H}=\{(S^t,z^t,\mathbf{X}^t)\}_{t=1}^T.$ Each instance in the dataset is a tuple where $t$ is a timestamp, $S^t$ is the set of stops served at $t,$ $\mathbf{X}^t$ denotes the actual preferred routing created by the planners, and $z^t$ are feature variables such as the demand of each stop, the number of vehicles used, the day of the week, or some other known parameters. Hereafter, we will use the symbols without the suffix $t$ to avoid notational complexity. 

\subsection{Transition Probabilities}

Explicit specification of the preferences of the drivers and the planners would result in a complex model with a large 
number of parameters to tune. Instead,
in this paper, we use the framework of \cite{canoy2019vehicle}, which captures the preferences of the route planners and the drivers using transition probabilities. In more formal terms, we learn a model which assigns probabilities to all the arcs within the network. Our hope is that these transition probabilities subsume the hidden preferences of the route planners and the drivers.
Formally, we learn $\Pr(r|s) $ which denotes the probability of the next stop being $r,$ conditional on the current stop $s$. We remark that the transition probability would be a function of some temporal and contextual attributes including but not limited to the traditional cost measures. 
\subsection{Maximum Likelihood Routing}
Once the probabilities are learned, we follow the methodology of \cite{canoy2019vehicle} to find the most \emph{likely} routing from the set of all feasible routings. We call the routing with the highest probability the maximum likelihood routing (MLE routing). Formally,
\begin{equation}
    \max_{x \in {\mathcal X}_S^m } \prod_{(s\rightarrow r) \in x} \Pr (r|s).
\end{equation}
In order to identify the MLE routing, we solve an optimization problem whose feasible region is defined by the following standard CVRP constraints \cite{toth2014family}. 
\begin{align}
    & \sum_{r\in V\!,\: r\neq s} x_{sr} = 1 && s\in S \label{eqn:con_flow1}\\
    & \sum_{s\in V\!,\: s\neq r} x_{sr} = 1 && r\in S \label{eqn:con_flow2}\\
    & \sum_{r=1}^n x_{0r} = m \label{eqn:con_fleet}\\
    & \text{if} \ x_{sr}=1 \ \Rightarrow \ u_s + q_r = u_r && (s,r) \in A : t\neq 0,\, s\neq 0 \label{eqn:con_cap1}\\
    & q_s \leq u_s \leq Q && s\in S\backslash\{0\} \label{eqn:con_cap2}\\
    & x_{sr} \in \{0, 1\} && (s,r) \label{eqn:con_integ}\in A.
\end{align}
\eqref{eqn:con_flow1} and \eqref{eqn:con_flow2} ensure that each customer is served by exactly one vehicle. \eqref{eqn:con_cap1} performs subtour elimination. \eqref{eqn:con_cap2} ensures that the vehicle capacity is respected. We remark that in \eqref{eqn:con_fleet}, we use the equality constraint because in practice, the company must use all the available vehicles.
The only modification from the standard CVRP is that instead of minimizing the distance, we maximize the joint probability. To transform the product in the objective function into a sum, we consider log probabilities in the objective function and minimize the following:
\begin{align}
\min_{x} & \sum_{(s,r)\,\in\, A} -log \Pr (r|s) x_{sr}
\end{align}

In the subsequent discussions, the $(s,r)$-th entry of matrix $P$ would contain $ \Pr (r|s)$.
\subsection{Transition Probability Estimation by Markov Counting}
\label{MarkovCounting}
The goal of the Markov Counting approach is to estimate all the conditional probabilities $\Pr(r|s)$ over the set of all stops in the data: $S^{\text{all}}=\bigcup_t S^t.$ From conditional probability theory, we have:
\begin{equation}
    \Pr(r|s) = \frac{\Pr(s\rightarrow r)}{\Pr(s)},
\end{equation}
where $\Pr(s)=\sum_u\Pr(s\rightarrow u).$ By defining the frequency of a transition $(s\rightarrow r)$ in the historical dataset $\mathcal{H}$ as $f_{sr} = \sum_t \llbracket\, (s\rightarrow r\in \mathbf{X}^t)\, \rrbracket
$, where $\llbracket\,\cdot\,\rrbracket$ equals 1 if the statement inside the bracket is true and 0 otherwise, the conditional probabilities from the dataset can be estimated by:
\begin{equation}
    \Pr(r|s) = \frac{f_{sr}}{\sum_u f_{su}}.
    \label{eq_markov_proba}
\end{equation}
We point out that with this formulation, we can solve the standard CVRP which minimizes the distance if we replace $\Pr(r|s)$ by a \emph{distance-based probability} $\text{Pr}_{dist}(r|s)$:
\begin{equation}
\text{Pr}_{dist}(r|s)
=  \frac{e^{-d_{sr}}}{\sum_u e^{-d_{su}}}.
 \label{eq:dist_proba}
\end{equation}

The transition probability matrix construction algorithm presented in \cite{canoy2019vehicle} makes use of \emph{weighing schemes,} where a variable weight $w_t$ is defined for each historical instance in $\mathcal{H}.$ This weight varies according to the properties of the tuple $(S^t,z^t,\mathbf{X}^t).$ Giving varying weights to each historical instance affects the way the transition frequencies are counted, hence each weighing scheme results in a different transition matrix. 
Exponential weighing is one of the most used scheme, where instances far-off in the past receive decaying weights. 
Therefore, in our experiments, we will compare our approach with the Markov model with the exponential weighing scheme.
\jay{Lize's comment: This paragraph seems a bit unexpected and lost at the end of this section, maybe you can explain the connection to the rest of the section a bit more (how a traditional CVRP can be seen as using probabilities as well)}

\section{Learning the Transition Probabilities Using Neural Network} \label{sect:methodology}
One limitation of the Markov counting approach introduced in section~\ref{MarkovCounting} is that it only uses past data to arrive at the probabilities. We want the transition probabilities to be a function of other attributes such as the day of week and the distances between stops, among others. This is the motivation behind using a neural network model.

Neural networks are made of interconnected units called neurons. A single neuron takes a series of inputs $z_0,...,z_n$ and returns an output $o$ as a function of the inputs $o =  f(\sum w_i z_i)$, or in matrix form $o = f(Wz)$, where $f$ is an activation function and $w_i$'s are the weights. Many choices for the activation function exist---\emph{sigmoid}, \emph{ReLU}, \emph{tanh} are some of the widely used activation functions. 
A network consists of several layers, and multiple neurons are stacked in each layer where the inputs are connected to each neuron. The output of the layer can be conveniently described in matrix form as $o = f(Wz)$. Here, each row of $W$ corresponds to each neuron. The dimension of output $o$ is controlled by the dimension of matrix $W$. In a multilayer network, the subsequent layers use the outputs of the preceding layers as inputs. Obviously, the designer has the option to transform the output between two layers.

A multilayer neural network is considered as a universal function approximator \cite{lin2014network}, which tries to learn the functional relationship between the output and the input. To do so, the parameters of the neural network must be learned using training data.
This is done by backpropagating the loss between the predicted output and the target output. During backpropagation, the derivative of the final loss with respect to the weights is computed and then the weights are updated by gradient descent. 
The choice of the loss function is dependent on the problem at hand. For a multiclass classification approach, categorical cross entropy loss is the preferred choice. 

We propose to learn the transition probabilities between the stops from the historical data using a neural network.
For a single day $t$, we have $(S^t,z^t, \mathbf{X}^t)$ as explained in section~\ref{MarkovCounting}. 
\subparagraph*{Feature variables.} We want the predicted probabilities to be a function of the feature variables. 
Different types of features can be considered: time-lagged temporal features, features related to the set of stops to be served ($S^t$), the distance between the stops and contextual features such as day of week, number of vehicles. The motivation behind using the time-lagged solutions as features is to learn from past solutions. We define the \emph{look back period} ($L$) as the maximum number of past observations considered in our model. The motivation of this look back period is two-fold: 1. it allows us to model the fact that past observations lose their relevance over time, and 2. to avoid problems of over-fitting due to lack of observations. 
We can also consider the output of the 
Markov counting model as a feature, as it subsumes past information. Moreover, this can be computed easily on the fly.
\begin{figure}[ht]
    \centering
    \includegraphics[scale=0.78]{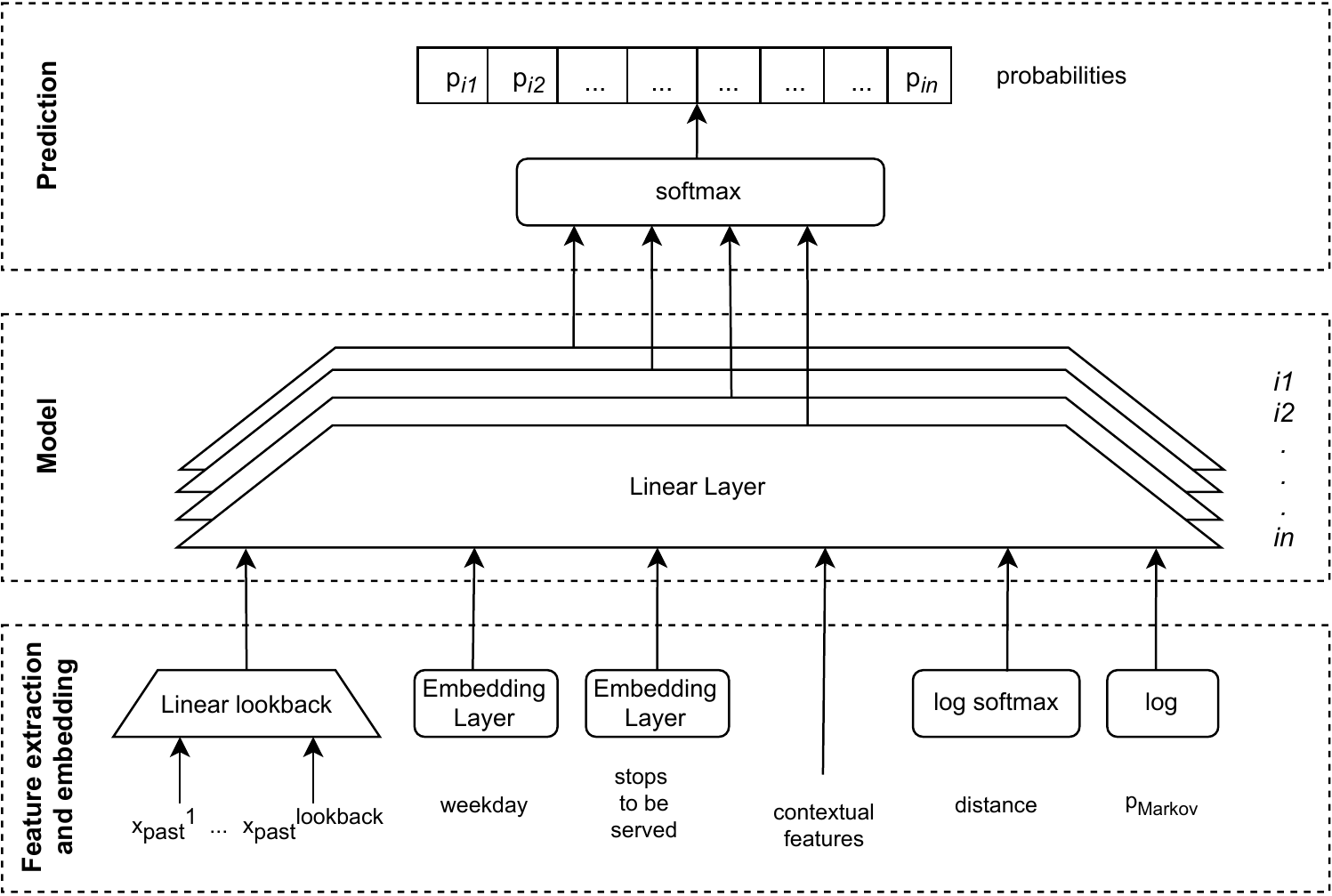}
    \caption{Neural network architecture to estimate transition probbailities from a source node $s$}
    \label{fig:neuralnet+markov}
\end{figure}

From this discussion, it is evident that some of the features are specific to an arc. This includes the time-lagged features, the distances and the Markov probabilities. On the other hand, features such as day of week and stops to be served are the same across all the stops. All of these are considered as an input to the neural network in Figure~\ref{fig:neuralnet+markov}.
\subparagraph{Architecture of the neural network.}
Our goal is to build a network that estimates the transition probabilities for each arc $(s,r) \in A$.
We have $|A| \times |A|$ distance features, $L \times |A| \times |A|$ time lagged features 
and $|K|$ number of contextual features. Because of limited data, our aim is to build a parsimonious network with as few parameters as possible.

Authors in~\cite{canoy2019vehicle} 
propose a linear combination of Markov probabilities (Eq.~\ref{eq_markov_proba}) and distance-based probabilities (Eq.~\ref{eq:dist_proba}), $
    P^t = \omega P^t_{Markov} + (1- \omega) P^t_{dist}.$
Essentially, this approach considers these two factors while computing the probabilities.
Furthermore, this can be extended to more number of feature variables in general. The advantage of using a neural network is that it learns the weights of the linear combination itself.

The final layer of the proposed architecture in Figure~\ref{fig:neuralnet+markov} performs this linear combination. 
This linear layer outputs unnormalized scores and a softmax operation upon these would result in the probabilities. As the outputs are unnormalized scores, so should be the inputs and that is why we log-transform the Markov probabilities. 
The distance based probabilities are arrived by considering the softmax of the distances from the source stop $s$.

We treat the categorical features such as day of week and stops to be served, by passing them to embedding layers before feeding them to the linear layer. The embedding layer computes dense vector representations of the categorical variables. The numerical feature variables such as the number of vehicles are passed directly to the final layer.

We treat the time-lagged solutions between $(s,r)$ by considering linear combinations of $L$ previous solutions of $(s,r)$. 
The first linear layer considers past solutions of the look back period as inputs and its output is a score based on that. Thus, the stop which has been chosen more often as the next stop, would be assigned a higher score.
We remark that there are other ways to treat time-lagged variables such as LSTM \cite{greff2016lstm}, but they might be prone to overfitting because they have a large number of parameters. 

The final linear layer outputs a score between $s$ and a destination node. So essentially, there are $|A|$ number of such linear layers. The use of neural network gives us the flexibility to use separate weights for the linear layers of every destination node. Obviously, this may result in overfitting as the number of parameters increase.
\tias{If space allows, please provide a more principled discussion of the features and how you model them here... Every 'block' in Fig1 needs an explanation using the same terminology. Also, I REALLY REALLY want the image to show that there are parallel linear blocks that merge into a softmax...}

Other than the contextual features, which are the same for all $(s,r)$, the other inputs to the linear layer are specific to $(s,r)$ and so is the output. We use a \textbf{separate model for each source node $s$} and each of them generates transition probabilities from the source node to all the other nodes. 
\begin{algorithm}[t]
\SetAlgoLined
\SetKwRepeat{Repeat}{repeat }{until convergence} 
\SetKw{Arg}{Parameter:}
\SetKwInOut{Input}{Input}\SetKwInOut{Output}{output}
\Input{\begin{itemize}
\item Historical solutions up to day T: $\{ (S^t,z^t, \mathbf{X}^t) \}_{t=1}^T$ \\
\item $s$: the source stop
\end{itemize}
}
active days $\longleftarrow$ list of days where $s$ is active \\
Model.initialize();\\
\For{$t$ in active days}
{
$x^t_{(s)} \leftarrow \mathbf{X}^t[s,:] $ \tcp{target solution }
$Past^t_{(s)}  \leftarrow  \{  \mathbf{X}^{t^\prime} [s,:]| t^\prime \in \text{active days[-(look back period):] }  \}$\\
\If{\text{length(active days) $<$ look back period}}{
Fill the remaining days with equiprobable probabilities vector\\
}
$\mathcal{H}^t \leftarrow \bigg(z^t_{(s)}, Past^t_{(s)} \bigg)$\\
\For{training epochs}{
$\hat{P}^t[s,:] \leftarrow \text{Model.Predict}(\mathcal{H}^t )$\\
$L \leftarrow \text{Cross Entropy}(\hat{P}^t[s,:], x^t_{(s)})$\\
Update Model by backpropagating $\nabla_{\hat{P}^t}(L )$\\
}}
 \caption{ Transition probability estimation from source stop $s$}
  \label{Algo:PredictiveModel}
\end{algorithm}

\subparagraph{Algorithm.} Algorithm~\ref{Algo:PredictiveModel} proposes a training scheme for estimating probabilities from a single source stop $s$ to the others.
We use all the features including the time-lagged solutions day $t$, to estimate the transition probabilities on day $t$. This past data ($\text{Past}_{(s)}^t$) is obtained by extracting the corresponding row from the incidence matrix.
As the training is for $s$, we only consider the next stop visited after it in the past. Hence, we formulate it as a multiclass classification problem, where the classes are the possible next stops and the objective is to classify them correctly. 
We use different models for different stops and while training the model for a stop $s$, we only consider past days where $s$ was served.
For any stop that does not have enough past data, to fill the look back period, the \textbf{remainder of the look back period are filled with uniform distribution} over the set of possible next stops. 
\subparagraph*{Loss function.} We formulate the learning problem as a multiclass classification task. The classification problem is to identify the next stop after $s$. 
While training we do not consider any VRP constraints, i.e., the transition probabilities of all the stops can be nonzero regardless of whether they are active or not. Once the neural network predicts the transition probability vector $\mathbf{p}_{(s)}$, we compute the cross entropy loss with respect to the actual solution $\mathbf{x}_{(s)}$.
\begin{align}
    L(\mathbf{p}_{(s)}, \mathbf{x}_{(s)}) = - \sum_{u \in V} x_{su} \log(p_{su}) 
\end{align}

Finally this loss is backpropagated to update the parameters of the neural network.
\begin{algorithm}[t]
\SetKwInOut{Input}{Input}
\SetKwBlock{Training}{begin training}{end training}
\SetKwBlock{Test}{begin evaluation}{end evaluation}

\Input{
$S^{T+1}, z^{T+1}, \{\mathbf{X}^t \}_{t=1}^{T+1}$
}
\For{each stop $\mathbf{s}$ in $S^{T+1}$}{
active days $\longleftarrow$ list of days $\mathbf{s}$ is active \\
Model $\leftarrow$ Algorithm~\ref{Algo:PredictiveModel} ($\{ (S^t,z^t, \mathbf{X}^t )\}_{t=1}^T$, $\mathbf{s}$) // Model training\\
\If{\text{length(active days) $\geq$ look back period}}{
$Past^T_{(s)}  \leftarrow  \{  \mathbf{X}^{T^\prime} [s,:]| T^\prime \in \text{active days[-(look back period):] }  \}$
}
\Else{
$Past^T_{(s)}  \leftarrow  \{  \mathbf{X}^{T^\prime} [s,:]| T^\prime \in \text{active days }  \}$ 
}
$\mathcal{H}^{T+1} \leftarrow \bigg(z^{T+1}_{(s)}, Past^{T+1}_{(s)} \bigg)$ \\
Model $\longleftarrow$ Model dictionary [$\mathbf{s}$] \\
$\hat{P}^{T+1}$[$\mathbf{s}$,:] $\longleftarrow$ Model.predict($\mathcal{H}^{T+1}$)
}
MLE Routing $(-log(P^{T+1}))$; Compare with $X^{T+1}$\\
\caption{Evaluation of Maximum Likelihood Routing}
\label{Algo:VRP}
\end{algorithm}
Algorithm~\ref{Algo:VRP} shows how we utilize the estimated transition probabilities to come up with the maximum likelihood solutions. Once we train the models for each stop using the data available until day T, we use it for routing on day $T+1$. To do so, first we estimate the transition probabilities for each stop using the trained models. 
The $(s,r)$-th entry of the matrix $\hat{P}$ contains the estimated transition probability of going from stop $s$ stop $r$.
Then using the estimated transition probability matrix $\hat{P}$, we solve the maximum likelihood routing problem.
\section{Decision Focused Learning}\label{sect:decisionfocused}
\begin{algorithm}[ht]
\SetAlgoLined
\SetKwRepeat{Repeat}{repeat }{until convergence} 
\SetKw{Arg}{Parameter:}
\SetKwInOut{Input}{Input}\SetKwInOut{Output}{output}
\Input{
Historic solutions till days T: $\{ (s^t,z^t, \mathbf{X}^t) \}_{t=1}^T$
}

\For{training epochs}{
\For{$t $ in 1 to $T$}{
\For{each stop $\mathbf{s}$ in $S^{T+1}$}{
$x^t_{(s)} \leftarrow \mathbf{X}^t[s,:] $ \\
$Past^t_{(s)}  \leftarrow  \{ \mathbf{X}^{t^\prime} [s,:]| t^\prime \in \textit{active days } \wedge |Past^t_{(s)}|\leq \text{lookback period} \}$\\
$\mathcal{H}^t \leftarrow \bigg(z^t_{(s)}, Past^t_{(s)} \bigg)$\\

$\hat{P}^t[s,:] \leftarrow \text{Model}^{(s)}{.Predict}(\mathcal{H}^t )$\\

}
$\hat{\pi} \leftarrow -log(\hat{P}^t)$\\
$\hat{X}^t \leftarrow \text{MLE Routing} (\pi)$\\
$L \leftarrow sum(ReLU(X^t -\hat{X}^t ))$\\
$\Tilde{\pi} \leftarrow \hat{\pi }- \lambda \frac{dL}{d \hat{X}^t}$\\
$\Tilde{X} \leftarrow \text{MLE Routing} (\Tilde{\pi})$\\
$\nabla_{\pi}(L )  \leftarrow -\frac{1}{\lambda} [\hat{X}^t - \Tilde{X} ]$\\
\For{each stop $\mathbf{s}$ in $S^{T+1}$}
{
$ \text{Model}^{(s)} \text{.backpropagate} \bigg(\nabla_{\pi}(L )[s,:] \bigg)$
}
}
}
 \caption{ Decision Focused Learning Algorithm }
 \label{Algo:Predict+Optimize()}
\end{algorithm}
The approaches proposed so far consider the prediction of the transition probabilities and the VRP optimization separately. 
Such approaches can be viewed as two-stage approaches \cite{demirovic2019investigation}, where a neural network model is separately trained to estimate the unknown coefficients of an optimization problem. 

One drawback of such a two-stage approach is the neural network model fails to incorporate information from the optimization problem.
As the neural network model is trained without regard for the downstream optimization problem, the loss function fails to consider the impact of the predicted coefficients on the final objective value of the optimization problem. 

Decision focused learning approaches \cite{elmachtoub2021smart,wilder2019}, on the other hand, consider how effective the predicted values are to solve the optimization problem and is trained with respect to the \emph{optimization task loss} rather than a prediction loss such as cross entropy loss.

In Algorithm~\ref{Algo:Predict+Optimize()} we show our implementation of the decision focused learning approach for this problem. We implement the methodology of \cite{Blackbox2020differentiation} to differentiate a combinatorial optimization problem with linear objective. \tias{This is still the old description? Or are we using this one indeed?} \jay{In terms of result, SPO and this approach has no difference,so I believe we can keep this}

They consider an optimization problem $\min_{X \in \chi} f(\pi,X)$ with a linear objective.
$\hat{X}^*(\hat{\pi})$ is the solution by using predicted $\hat{\pi}$ and  the final optimization task loss is $L(X^*(\pi),\hat{X}^*(\hat{\pi}))$.   The gradient of this task loss with respect to $\hat{\pi}$ is the following
\begin{align}
    \nabla_{\hat{\pi}} L(X^*(\pi),\hat{X}^*(\hat{\pi})) = -\frac{1}{\lambda} [\hat{X}^*(\hat{\pi}) - \tilde{X}^*(\tilde{\pi})]
\end{align}
where $\tilde{\pi}$ is a perturbation around the predicted $\hat{\pi}$, given by
\begin{align}
    \tilde{\pi} =  \hat{\pi} + \lambda \frac{d L(X^*(\pi),\hat{X}^*(\hat{\pi}))}{d \hat{X}^*(\hat{\pi})}
\end{align}

In our setting, $\pi$ is the matrix of negative log probability vectors i.e. $\pi = - \log (P)$, and $X$ is the resulted routing. The final task is to minimize the difference between the actual route and the proposed route. So a suitable choice for the \emph{task loss} is to consider arc difference, the number of arcs present in the actual solution but not in the predicted solution.  Formally,
\begin{align}
    L(X^*(\pi),\hat{X}^*(\hat{\pi})) = Sum( ReLU(X^*(\pi) - \hat{X}^*(\hat{\pi}))) = \sum_{(i,j) \in \text{dim}(X)} \max \big( x_{ij}- \hat{x}_{ij},0 \big)
\end{align}
here $Sum$ is the summation of all the elements of the matrix. The derivative of $L$ can be computed s follows
\begin{align}
    \frac{d L(X^*(\pi),\hat{X}^*(\hat{\pi}))}{d \hat{X}^*(\hat{\pi})} = \begin{cases}
                        -1 \text{  if  } \hat{x}_{ij} < x_{ij} \\
                        0 \text{  otherwise  }
                    \end{cases}
                    \forall (i,j) \in \text{dim}(\hat{X}^*)
    \label{eq:decision_grad}
\end{align}
If we consider a squared loss instead of the ReLU, we replace $\max \big( x_{ij}- \hat{x}_{ij},0 \big)$ with $ \big( x_{ij}- \hat{x}_{ij} \big)^2$. In this case the derivative would be
\begin{align}
    \frac{d L(X^*(\pi),\hat{X}^*(\hat{\pi}))}{d \hat{X}^*(\hat{\pi})} = \begin{cases}
                        -2 \text{  if  } \hat{x}_{ij} < x_{ij} \\
                        2 \text{  if  } \hat{x}_{ij} > x_{ij} \\
                        0 \text{  otherwise  }
                    \end{cases}
                    \forall (i,j) \in \text{dim}(\hat{X}^*)
\end{align}
Intuitively, if $i\rightarrow j$ is present in $X$, but not in $\hat{X}$, then we lower $\pi_{ij}$ by $\lambda$ and with this generate a new solution with. A scaled difference between these two solutions is the gradient with respect to $\pi$.

\section{Experimental Evaluation }
\subsection{Evaluation Criteria}
We are interested in how the MLE routing solutions differ from the used routes. 
To do so, we evaluate the performance using the following two evaluation measures.
\subparagraph{Arc Difference (AD)} measures the number of  arcs traveled in the actual solution but not in the MLE routing solution. It is calculated by taking the set difference of the arc sets of the test and predicted solutions. The percentage is computed by dividing AD by the total number of arcs in the whole routing.
\subparagraph{Route Difference (RD)} indicates the number of stops that were incorrectly assigned to a different route. Intuitively, RD may be interpreted as an estimate of how many moves between routes are necessary when modifying the predicted MLE solution to match the actual routing. To compute RD, the pair of routes with the smallest difference in stops is greedily selected without replacement. The total number of incorrectly assigned stops is considered as RD.  The percentage is computed by dividing RD by the total number of stops in the whole routing.

We also present the cross entropy (CE) loss as the neural network models are trained with respect to this criterion.
\subsection{Data Description}

For empirical evaluation\footnote{The code and the anonymised data are available at https://github.com/JayMan91/CP2021-Data-Driven-VRP}, we use actual historical data from a logistics company to compare the performance of our proposed approaches against the Markov model presented in \cite{canoy2019vehicle}. The data consists of 201 daily routings collected in a span of 39 weeks. It has 73 unique customers, each representing a node other than the depot. In each instance, an average of 31 stops are serviced by an average of 8 vehicles. We group the instances by day of the week, giving us an average of 29 instances per weekday. In training and testing the models, we used a 75\%-25\% split while ensuring that we have exactly 7 testing instances per weekday. 
We use a rolling window model for valuation, where the lookback period remains fixed and counts backwards from the most recent observation.

In Table~\ref{tab:markovnnweekly}, we present the percentage AD and RD of the Markov approaches on the test instances for each day of week.  The Markov (allday) approach arrives on the transition probability by considering all past days. On the contrary, the Markov (weekday) approach considers only those past instances which occurred on the same day of week as the evaluation instances. Both the approaches use Eq.~\eqref{eq_markov_proba} to compute the probabilities. 
We can see in Table~\ref{tab:markovnnweekly} that Markov (allday) performs better on the weekdays, but its result worsens on the weekends. 
 Due to the operational characteristics of the company, the number of customers, number of available vehicles, and hence the routing decisions tend to be highly dependent on the day-of-week. So 
there is a strong influence of the day on the probabilities. Probably, this is why \cite{canoy2019vehicle} preferred the Markov (weekday) approach.
The motivation behind the neural network approach is that it can consider the day of week as feature, so that we do not have to compute the probability separately for each day. Moreover, other feature variables can easily be passed into the neural network.

In our experiment we consider the following feature variables to predict the transition probabilities$-$ a. day of week, b. the set of stops to be served, c. distance between the stops, d. number of available vehicles, e. routings used in the past, f. transition probabilities computed by Markov (weekday).
\subsection{Experimental  Results}
In this section, we will address the following research questions
  \begin{itemize}
    \item Choice of the feature variables and the network architecture in a systematic way
      \item Compare the quality of predictions of the neural network trained with respect to the CE loss with that of Markov counting approach
      \item The effectiveness of a decision focused approach, which trains the network to directly minimize AD
  \end{itemize}
\subsubsection{ Choice of Network Architecture and Feature Variables }
As mentioned in Section~\ref{sect:methodology}, there are many choices for the network architecture and because of limited data we are careful to avoid overfitting.\footnote{We use  Pytorch \cite{Pytorch} and Gurobi \cite{gurobi} for neural network and VRP models respectively.}
In Table~\ref{tab:ablation}, we first present the impact of feature variables on the quality of predictions. We show the cross entropy loss on both training and test data and AD and RD on test data.
We point out that the network presented in Figure~\ref{fig:neuralnet+markov} results in the lowest training loss among them.  On the other hand, a network only with Markov probabilities as input lowers test loss and lower AD, RD. 
A network without the time-lagged data has even lower CE loss on the test instances and lowest AD and RD suggesting that past information is already subsumed in the Markov probabilities, making the time-lagged information redundant. The Markov probabilities along with the contextual information seem to be the right choice for the feature variables.
\begin{table}[t]
    \centering
    \begin{tabular}{lrrrrrr}
\toprule
                                 Model & Training CE &  Test CE &    AD &  AD (\%) &    RD &  RD(\%) \\
\toprule
\multicolumn{7}{c}{Experiment on feature variables}\\
\midrule
Neural Net  &          \textbf{ 2.14} &     1.10 &  6.27 &   19.80 &  4.57 &  18.04 \\
         \makecell[l]{Neural Net \\(without past data)} &           2.48 &    \textbf{ 1.04} & \textbf{ 5.68 }& \textbf{  18.04} &  \textbf{ 4.30} & \textbf{  17.02} \\
        \makecell[l]{Neural Net \\(without weekday)}  &          \textbf{ 2.14} &     1.09 &  6.24 &   19.75 &  4.59 &  18.03 \\
  \makecell[l]{Neural Net \\(without stop information)}   &           2.20 &     1.13 &  6.28 &   19.86 &  4.56 &  17.97 \\
  \makecell[l]{Neural Net \\(without distance)} &           2.20 &     1.10 &  6.18 &   19.54 &  4.46 &  17.62 \\
 \makecell[l]{Neural Net \\(without Markov probabilities)} &           2.43 &     1.49 &  7.99 &   26.26 &  5.32 &  21.38 \\
  \makecell[l]{Neural Net \\(only Markov probabilities)}  &           2.58 &     1.07 &  5.95 &   18.85 &  \textbf{4.29} &  \textbf{16.93} \\
\bottomrule
\multicolumn{7}{c}{Experiment on architecture choice}\\
\midrule
    \makecell[l]{LSTM } &           2.22 &    \textbf{ 1.01} &  6.35 &   20.10 &  4.49 &  17.75 \\
    \makecell[l]{Linear Layer \\ different for each stop} &           \textbf{1.37} &     1.82 &  7.21 &   22.81 &  4.74 &  18.57 \\
\bottomrule
\end{tabular}
    \caption{Investigation into feature variables and architectures}
    \label{tab:ablation}
\end{table}

In the lower section of Table \ref{tab:ablation}, we present two alternative architecture choices. The first one replaces the linear layer of the lagged solutions with an LSTM. The second one has different weights for different destination stops in the final layer in  Figure~\ref{fig:neuralnet+markov}.

It shows that using different weights for different destination stops results in lowest training CE loss. But this model clearly overfits, as the performance is poor on the test data.  The LSTM model seems to improve on CE loss but not on AD and RD measures. 
So overall, the model \textbf{without the past data} results in lowest CE loss as well as lowest AD and RD. This model has the Markov probabilities as input, and the past information carried by this probability.
\subsubsection{Neural Network Predictions}
\begin{table}[ht]
    \centering
    \begin{tabular}{llrllll}
\toprule
 {} & CE loss & \multicolumn{2}{c}{Arc Difference (AD)} &  \multicolumn{2}{c}{Route Difference (RD) } & \makecell{Distance\\(km.)}\\
\midrule
 &  &     Absolute &    Percent &             Absolute &    Percent \\
\midrule
  \makecell[l]{Markov \\ (allday)}                              &     2.77 &                         10.33 &    35.69 &       6.29 &       25.75 &                424 \\
 \makecell[l]{Markov \\ (weekday)}                           &     2.44 &                          5.86 &    18.55 &    4.39 &        17.26 &                  418 \\
 \makecell[l]{Neural \\ Net} &     \textbf{1.04} &           \textbf{ 5.68} &    \textbf{18.04} &                          \textbf{  4.30} &    \textbf{17.02} &  414 \\
 \makecell[l]{Conventional \\VRP}                     &    11.90 &                       21.47 &                     73.14&          11.65 &   46.93 &  \textbf{366 }\\
\bottomrule
\end{tabular}
    \caption{Comparison of Neural Network with Markov Counting (Actual Distance is 413 km.) }
    \label{tab:neural+markov}
\end{table}
The last section suggests to consider a network without the lagged variables for this task.
Next, we compare the quality of predictions of  this model shown to that of Markov counting approach. We present the average of CE loss, AD, RD between the actual solutions and generated solution on test instances in Table~\ref{tab:neural+markov}. We also present the distance of the solutions of these approaches.
We point out in Table~\ref{tab:neural+markov} that a neural network model results in lower CE loss, which is expected as the model is trained with that objective.
Moreover, we also observe lower AD and RD with this model. 
We also present results of a conventional VRP algorithm, which is the best in terms of total distance covered, but clearly very far off from the preferred solution.
\begin{table}[ht]
    \centering
    \begin{tabular}{lrrr|rrrrr}
\toprule
{} & \multicolumn{3}{c}{Arc Difference(\%)} & \multicolumn{3}{c}{Route Difference(\%)} \\
\toprule
{} &  \makecell{Markov\\ (allday)} &  \makecell{Markov\\ (weekday)} &  \makecell{Neural\\ Net} & \makecell{Markov\\ (allday)} &  \makecell{Markov\\ (weekday)} &  \makecell{Neural\\ Net} \\
\toprule
Monday &             51.53 &        \textbf{23.62} &                            \textbf{23.62} &               27.98 &        21.75 &                               \textbf{ 21.25 }\\
Tuesday &            \textbf{ 24.96 }&        25.61 &                               25.82 &               29.15 &        \textbf{28.87} &                                30.85 \\
Wednesday &            \textbf{ 19.61 }&        21.30 &                               20.48 &               17.96 &        15.12 &                                \textbf{14.61 }\\
Thursday &             24.89 &        22.86 &                               \textbf{ 21.17 }&               19.75 &       18.63 &                                \textbf{17.08} \\
Friday &             19.18 &        19.38 &                             \textbf{ 18.08} &               13.74 &       13.19&                                \textbf{11.54} \\
Saturday &             51.59 &       \textbf{  0.00} &  \textbf{  0.00} &               25.21 &        \textbf{  0.00} &                                 \textbf{  0.00} \\
Sunday &             58.09 &       \textbf{ 17.08} &                              17.11 &               46.48 &       \textbf{ 23.23 }&                                23.82 \\
\bottomrule
Overall &  35.69 	 & 18.55 &  \textbf{18.04} & 25.75 & 17.26 & \textbf{17.02} \\
\bottomrule
\end{tabular}
    \caption{Daywise Analysis of Arc Difference and Route Difference}
    \label{tab:markovnnweekly}
\end{table}
Table~\ref{tab:markovnnweekly} presents this comparison in more detail, where we evaluate for each day of week separately. Although we do not need to train the neural network separately for each day of week, by considering the Markov probabilities as inputs, it is able to generate 
predictions which result in lower AD and RD.
This demonstrates the advantage of the neural network approach,  which can consider multiple inputs, contextual as well as temporal, in a single framework.

\begin{figure}[t]
    \centering
    \begin{minipage}{0.260\textwidth}
        \centering
        \includegraphics[width=\linewidth]{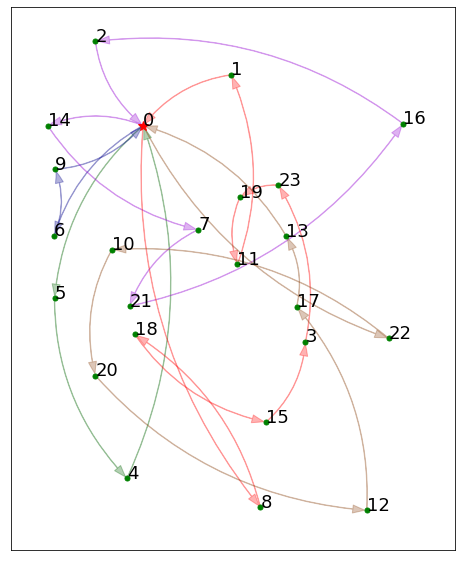}
        \caption{Human-made solution \\ \phantom{}}
        \label{fig:map_dist}
    \end{minipage}%
    \begin{minipage}{0.01\textwidth}
        \hspace{0.1cm}
    \end{minipage}%
    \begin{minipage}{0.305\textwidth}
        \centering
        \includegraphics[width=\linewidth]{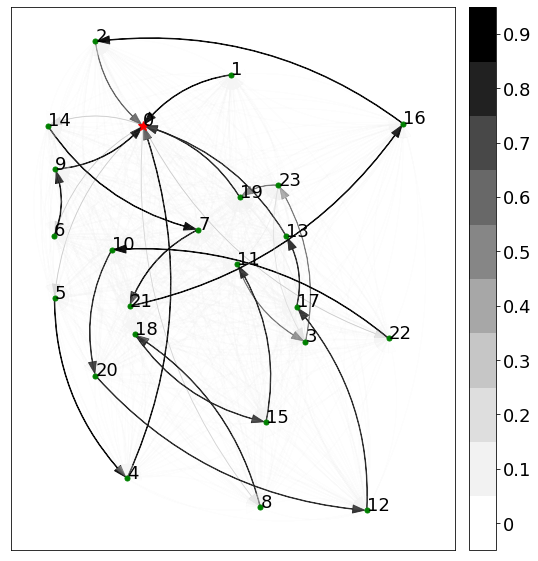}
        \caption{Learned proba-\\bilities (Markov)\\ \phantom{}}
        \label{fig:map_sol}
    \end{minipage}
    \begin{minipage}{0.01\textwidth}
        \hspace{0.05cm}
    \end{minipage}%
    \begin{minipage}{0.305\textwidth}
        \centering
        \includegraphics[width=\linewidth]{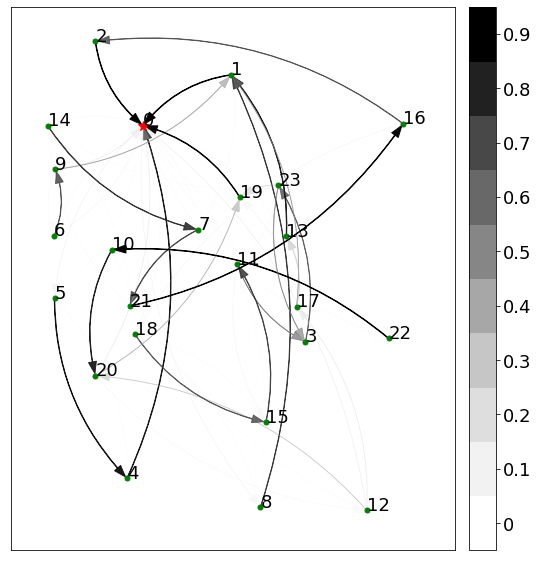}
        \caption{Learned proba-\\bilities (NN)\\ \phantom{}}
        \label{fig:map_trans}
    \end{minipage}

$ $

$ $

    \centering
    \begin{minipage}{0.260\textwidth}
        \centering
        \includegraphics[width=\linewidth]{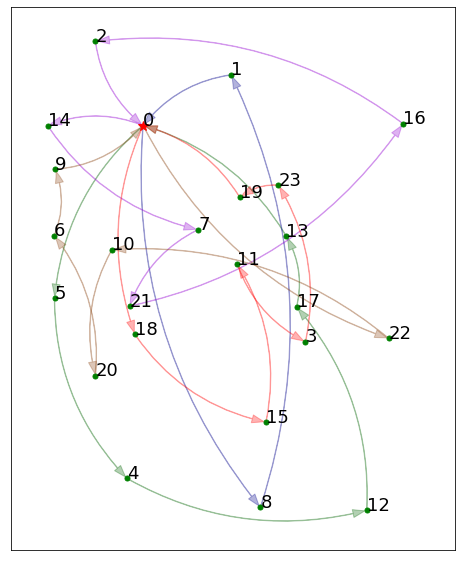}
        \caption{Markov\\ solution}
        \label{fig:map_pred}
    \end{minipage}%
    \begin{minipage}{0.075\textwidth}
        \hspace{1.2cm}
    \end{minipage}%
    \begin{minipage}{0.260\textwidth}
        \centering
        \includegraphics[width=\linewidth]{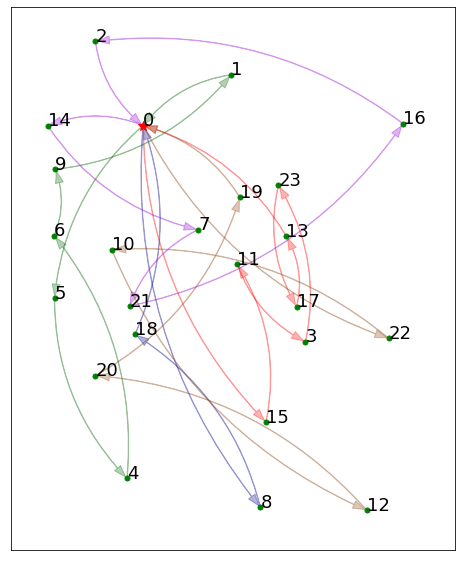}
        \caption{NN\\ solution}
        \label{fig:map_pred_2mm}        
    \end{minipage}
\end{figure}

Figures~\ref{fig:map_dist} to \ref{fig:map_pred_2mm} illustrate our approach for one instance.
Figures~\ref{fig:map_sol} and \ref{fig:map_trans} present the learned transition probabilities and Figures~\ref{fig:map_pred} and \ref{fig:map_pred_2mm} show the MLE routing of Markov weekday  and neural net respectively.
\subsubsection{Decision Focused Learning}
Next, we experiment with the decision focused learning approach introduced in section~\ref{sect:decisionfocused}. 
 We use the same neural network architecture but trained with arc difference as the loss function, and the loss backpropagted through a corresponding subgradient (Eq.~\eqref{eq:decision_grad}).
\begin{table}[ht]
    \centering
    \begin{tabular}{llrllll}
\toprule
 {} & \makecell{CE loss \\(test)} & \multicolumn{2}{c}{Arc Difference (AD)} &  \multicolumn{2}{c}{Route Difference (RD) } & \makecell{Distance\\(km.)}\\
\midrule
 &  &     Absolute &    Percent &             Absolute &    Percent \\
\midrule
  \makecell[l]{Relu \\loss }                              &     2.77 &                         13.76 &    45.96 &       9.51 &       38.54&                434 \\
 \makecell[l]{Squared  \\ loss}                           &     3.97 &                          13.31 &    44.27 &    9.10 &        37.14 &                  436 \\
\bottomrule
\end{tabular}
   \caption{AD and RD with Decision Focused Learning}
    \label{tab:Pred+Opt}
\end{table}
We present the solution quality of this approach in Table~\ref{tab:Pred+Opt}.  We can see, it fails to generate lower AD and RD on the test instances.
In fact, we observe AD reducing on training instances but not on test instances, suggesting a case of overfitting. 
Only 152 instances is not enough to train a complex model like this.
So the poor quality can be attributed to limited amount of data. 
\section{Conclusion}
We presented a neural network model which learns the transition probabilities between stops in a CVRP setting. With these transition probabilities, we solve the MLE routing problem instead of the conventional VRP. The resulting solution is able to mimic the solution preferred by the route planners and drivers. In this way, we are able to include the preferences of the planners and the drivers implicitly in the VRP solution.

We extend on the work of \cite{canoy2019vehicle}. The novelty of our approach is to use a neural network model to estimate the probabilities. Key developments are the use of an arc-based architecture to control the number of trainable parameters, and the identification of the standard cross-entropy classification loss as a suitable (and cheap to compute) proxy loss for training. This leads us to develop a general framework for such problem setting, which has the flexibility of taking contextual features including the output of \cite{canoy2019vehicle} into consideration.
By considering the contextual features in a principled way, our approach marginally outperforms \cite{canoy2019vehicle}, emphasising the advantage of a generic approach. 

We also use a decision focused learning approach which directly trains the neural network to minimize the final objective of minimizing the difference between the generated solution and the preferred solution. Although this approach considers the structure of the VRP optimization problem, our results show that it fails to generate good quality solutions in the test data. We believe it is due to the limited number of training instances. 

Our methodology relies on the presence of recurrent stops in our training set. Future research will aim to extend our methodology to learn preferences over non-recurring stops. It will also be interesting to investigate loss functions that include the structure of the CVRP tackling the challenges of the scalability. 


\bibliography{mybibliography.bib}
\newpage
\appendix
\section{Experiment Setup}
We use  Pytorch \cite{Pytorch} and Gurobi \cite{gurobi} for neural network and VRP models respectively. We use Adam optimizer~\cite{KingmaB14} implementation of Pytorch. The hyperparameters of each setup is detailed below.
\begin{table*}[ht]
\centering
\begin{tabular}{lcccccccccc}
\toprule
&  \makecell{Learning \\rate} & Epochs \\
 \midrule
Neural Net  &   0.1 & 50      \\
         \makecell[l]{Neural Net \\(without past data)} & 0.1 & 100 \\
        \makecell[l]{Neural Net \\(without weekday)} & 0.1 & 50  \\
  \makecell[l]{Neural Net \\(without stop information)}  & 0.1 & 100  \\
  \makecell[l]{Neural Net \\(without distance)} & 0.1 & 100\\
 \makecell[l]{Neural Net \\(without Markov probabilities)} & 0.1 & 100 \\
  \makecell[l]{Neural Net \\(only Markov probabilities)} & 0.1 & 100 \\
  \makecell[l]{LSTM } & 0.1 & 50 \\
    \makecell[l]{Linear Layer \\ different for stops} & 0.01 & 100 \\
\bottomrule
\end{tabular}
\caption{Hyperparameters Configuration (For all experiments the embedding dimension of weekday and stop feature are 6 and 40 respectively)}
\end{table*}

\end{document}